\documentclass[letterpaper, 10 pt, conference]{ieeeconf}  

\IEEEoverridecommandlockouts                              

\usepackage[acronym,nonumberlist]{glossaries}
\usepackage{amsmath}
\usepackage{amssymb}
\usepackage{xfrac}
\usepackage{mathtools}
\usepackage{relsize}

\usepackage{xspace}
\usepackage{hyperref}
\usepackage{soul}

\usepackage{array}
\usepackage{threeparttable}
\usepackage{tabularx}
\usepackage{multirow}

\usepackage{cleveref}

\usepackage[font=footnotesize]{subcaption}

\usepackage[acronym]{glossaries}
\newacronym{nerf}{NeRF}{Neural Radiance Fields}
\newacronym{ict}{ICT}{Information and Communications Technology}
\newacronym{ibvs}{IBVS}{Image-Based Visual Servoing}
\newacronym{sam}{SAM}{Segment Anything Model}
\newacronym{sfm}{SFM}{Structure-from-Motion}
\newacronym{ccr}{CCR}{Cross-Correlation Ratio}

\begin{document}

\title{
\Large
\bf
NeRFoot: Robot-Footprint Estimation for Image-Based Visual Servoing
}
\author{Daoxin Zhong, Luke Robinson and Daniele De Martini
\\
Mobile Robotics Group (MRG), University of Oxford\\\texttt{\{daoxin.zhong,luke.robinson,daniele\}@oxfordrobotics.institute}
\thanks{
This work was supported by EPSRC Impact Acceleration Account (IAA) ``Robotics  Inversion''.
}
}
\maketitle

\maketitle

\begin{abstract}
This paper investigates the utility of \gls{nerf} models in extending the regions of operation of a mobile robot, controlled by \gls{ibvs} via static CCTV cameras. Using \gls{nerf} as a 3D-representation prior, the robot's footprint may be extrapolated geometrically and used to train a CNN-based network to extract it online from the robot's appearance alone.
The resulting footprint results in a tighter bound than a robot-wide bounding box, allowing the robot's controller to prescribe more optimal trajectories and expand its safe operational floor area.


\end{abstract}

\glsresetall

\section{Introduction}
Visual servoing is a robotics technique that provides control based on visual feedback from external cameras.
Since \cite{hill1979real}, the field has evolved to encompass various methodologies and approaches.
Here we focus on \gls{ibvs} \cite{IBVS_1,IBVS_2}, which is typically about the specific 3D information of the robot and scene and performs state estimation and control of the robot in the image space.

\begin{figure}
    \centering
    \includegraphics[width=0.30\textwidth]{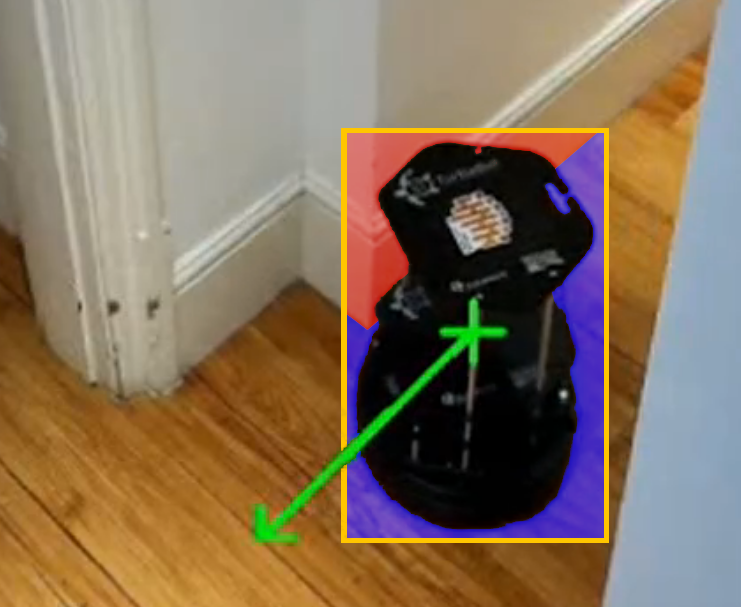}
    \caption{\cite{robinson2023robot} controls the robot based on its bounding box (yellow) and orientation (green). When checking if a trajectory is safe, its box must stay within the drivable region (blue). However, this precludes a huge area, which is still safe but intersects with the wall (red) in the image plane.\label{fig:problem}}
    \vspace{-15pt}
\end{figure}

Following \cite{robinson2023robot,robinson2023visual}, we aim for a generalised pipeline for controlling a mobile robot indoors (e.g., a restaurant or a warehouse).
The mobile robot is designed to be a simple, drive-by-wire apparatus capable of only receiving actuator commands from a \gls{ict} infrastructure.
The robot's position is captured via various CCTV cameras placed within the environment and is processed to supply a control signal to the mobile robot, directing it towards some navigational target.
Critically, no 3D information about the robot or the environment is supplied to the system, leading to suboptimal use of the driving surfaces, as shown in \cref{fig:problem}.

Drawing inspiration from previous works leveraging CAD as known prior \cite{CAD}, we overcome these issues by first building a \gls{nerf}-based \cite{vanilla-nerf} 3D representation of the robot using images captured from our CCTV cameras.
The robot's footprint, i.e. its contour when projected into the ground plane, is then estimated using an image of the robot synthesised from a downward perspective.
Its outline is then reprojected into the camera plane of the original images to generate masks and train a YOLO \cite{Yolo} segmenter for online footprint estimation at operation time.

\section{Methodology}

Our proposed methodology consists of two sequential steps: \gls{nerf} model training, and YOLO footprint estimation.

\subsection{NeRF Model Training}
We start from the same calibration data collected from \cite{robinson2023visual}, consisting of a video of the robot spinning in place with its bounding boxes. 
To train a \gls{nerf} model, a mask for the robot is first generated \textit{at every frame} from \gls{sam} \cite{sam} by seeding it with the bounding boxes.
This mask serves two purposes.
It segments the robot from the static background, allowing the relative pose of the camera to the robot to be estimated via traditional \gls{sfm} methods or more recent transformer models such as DUSt3R\cite{wang2024dust3r}.
Additionally, it selects the relevant ray bundles to be passed onto Nerfstudio \cite{nerfstudio} that informs the \gls{nerf} about the colour and density functions of the robot.
Critically, the iNeRF \cite{inerf} correction mechanism helps correct the estimated camera poses and improves the overall render quality.
This is further improved by optimising the camera poses along a plane, as constrained by the physical dimensions of the experimental setup.


\subsection{YOLO Footprint Estimation}
\label{sec:met:footprint}

We first synthesise a downward view of the robot and extract its mask via \gls{sam} \cite{sam}.
The coordinates of the footprint contours in the world frame are calculated by solving for the points of intersection between the projection rays and the ground plane.

To estimate the robot's footprint during operation, we train a segmentation network using a synthetic dataset of the robot rendered via \gls{nerf}.
By projecting the footprint contours into the image plane, we generate a corresponding footprint mask that varies with the robot's appearance.
This is used to train a YOLOv8 segmentation network, which is fast enough to provide real-time segmentation during deployment.
\Cref{fig:footprint_pipeline} shows examples of the footprint taken from a CAD model and the estimates from YOLO trained on the CAD or the synthetic dataset, showing how YOLO can segment the footprint reliably from the constructed data.

\section{Experimental Results}

Here, we report the results obtained in simulation on a Clearpath Jackal robot model using Blender, where we apply the same calibration approach as in \cite{robinson2023visual}.

\subsection{Synthetic Image Generation}
\label{sec: synthetic Image gen}

\gls{nerf} is notoriously data-hungry.
As such, we first analyse how many data points we need to create a usable render in \cref{tab:sprse_metrics}. 
Unsurprisingly, we notice how more images produce a better model for rendering; however, this incremental improvement is not uniform, and it becomes marginal after 150 images.
In a qualitative assessment, 166 of 200 generated images are deemed sufficiently photorealistic, closely resembling their counterparts in the original dataset.

Similarly, as the synthetic camera positions will be inherently noisy, we show ablations on an initial error in \gls{nerf}'s rendering quality.
The results are reported in \cref{tab:jackal_error}.
Here, we analyse the utility of our constrained camera-pose optimisation.
We can see how ours (plane opt.) always outperforms the default optimizer.

\begin{table}
    \centering
    \begin{tabular}{c|c|ccc}
         \# Spins & Frames ps & PSNR ($\uparrow$) & SSIM ($\uparrow$) & LPIPS ($\downarrow$) \\
        \hline
        \multirow{3}*{6} & 50 & 41.0 & 0.993 & 0.00256\\
         & 25 & 43.2 & 0.995 & 0.00188\\
         & 10 & 34.0 & 0.985 & 0.01556\\
        \hline
        \multirow{3}*{3} & 50 & 40.4 & 0.993 & 0.00479 \\
         & 25 & 33.6 & 0.985 & 0.01595 \\
         & 10 & 33.3 & 0.984 & 0.01709\\
        \hline
        \multirow{3}*{1} & 50 & 33.6 & 0.984 & 0.01714 \\
         & 25 & 33.5 & 0.984 & 0.01708 \\
         & 10 & 33.3 & 0.984 & 0.01672\\

    \end{tabular}
    \caption{Average metrics by spins count and number of sampled images per spin.}
    \label{tab:sprse_metrics}
\end{table}

\begin{table}
    \centering
    \begin{tabular}{l|c|ccc}
         & Avg Error & PSNR ($\uparrow$) & SSIM ($\uparrow$) & LPIPS ($\downarrow$) \\
        \hline
        Baseline & 0 & 56.0& 0.999 & 0.00037\\
        \hline
        \multirow{4}*{Default Opt.} & 1  &34.8 & 0.987 &0.0135\\
         & 2  & 34.8& 0.987 & 0.0139\\
         & 5  & 34.3& 0.986 &0.0151\\
         & 10  & 34.2 &0.985 &  0.0164 \\
        \hline
        \multirow{4}*{Plane Opt.} & 1  &41.0 & 0.993 & 0.00256 \\
         & 2  &41.8 &0.994 & 0.00259 \\
         & 5  &34.4 &0.986 & 0.00152 \\
         & 10  &34.2 & 0.985&0.00165  \\
    \end{tabular}
    \caption{Average metrics by varying error magnitude.}
    \label{tab:jackal_error}
    \vspace{-10pt}
\end{table}

\begin{figure}[t]
    \centering
    \begin{subfigure}{0.47\textwidth}
    \centering
    \includegraphics[width=0.8\textwidth]{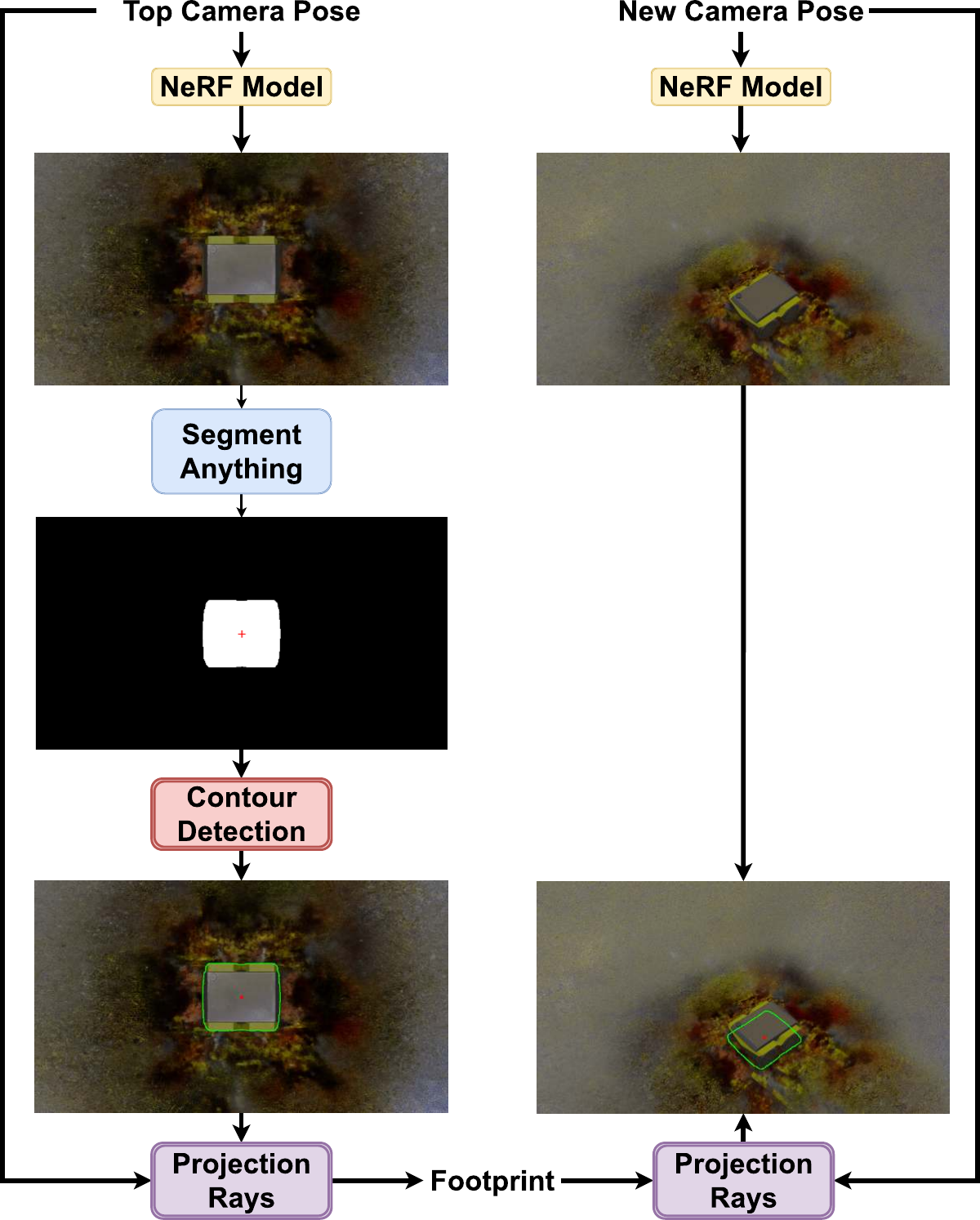}
    \vspace{5pt}
    \end{subfigure}
    \begin{subfigure}{0.45\textwidth}
        \centering
    \begin{subfigure}{0.32\textwidth}
        \includegraphics[width=\textwidth,trim=400 120 450 300, clip]{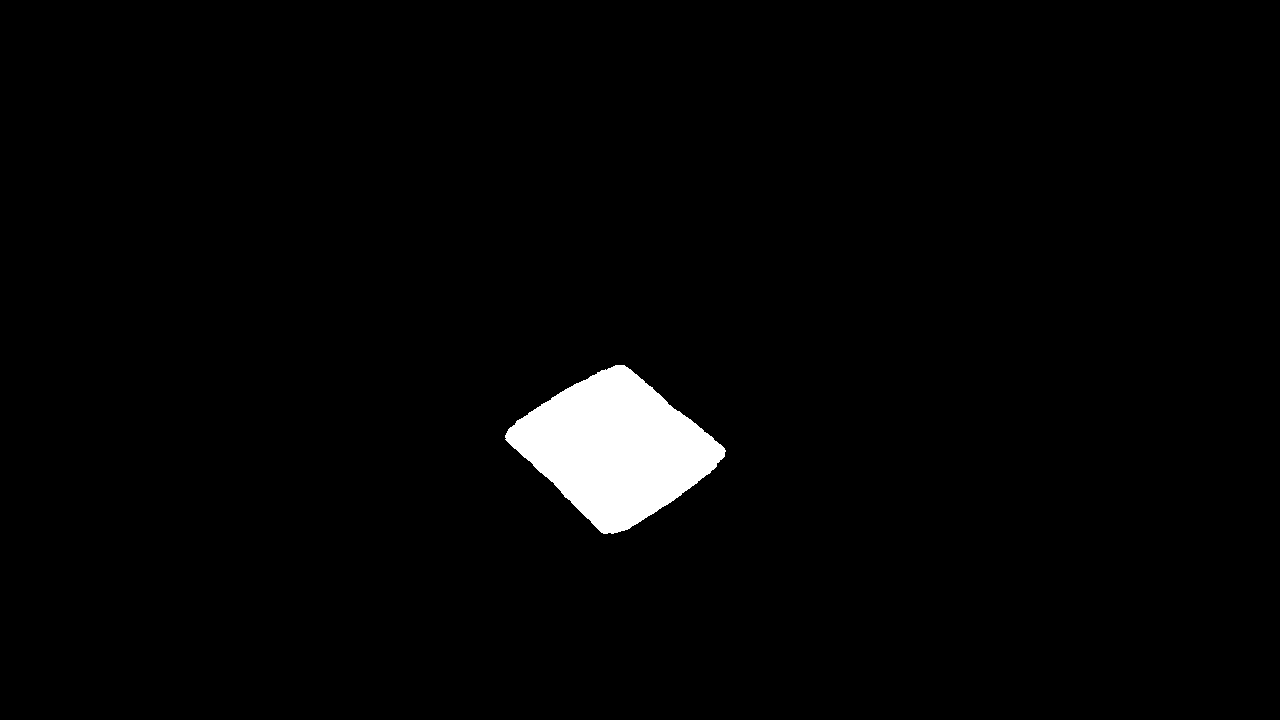}
        \caption{\label{fig:gt_foot}}
    \end{subfigure}
    \begin{subfigure}{0.32\textwidth}
        \centering
        \includegraphics[width=\textwidth,trim=400 120 450 300, clip]{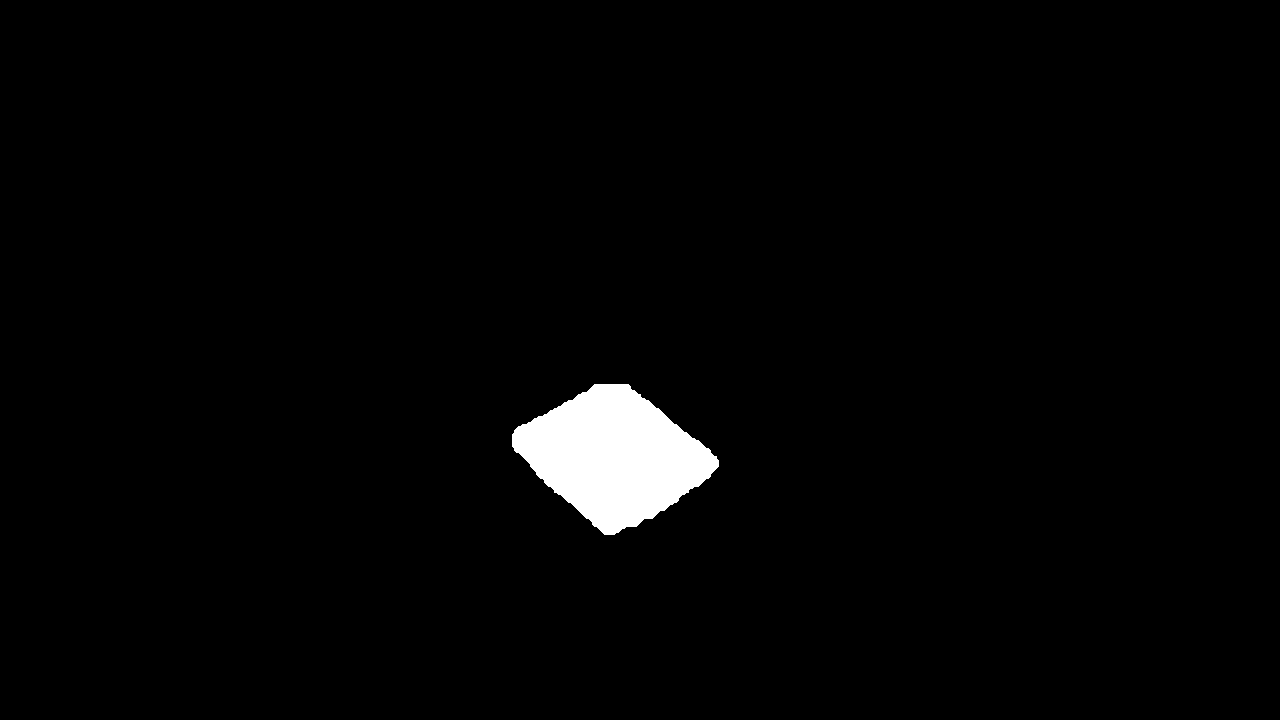}
        \caption{\label{fig:gt_foot_yolo}}
    \end{subfigure}
    \begin{subfigure}{0.32\textwidth}
        \centering
        \includegraphics[width=\textwidth,trim=400 120 450 300, clip]{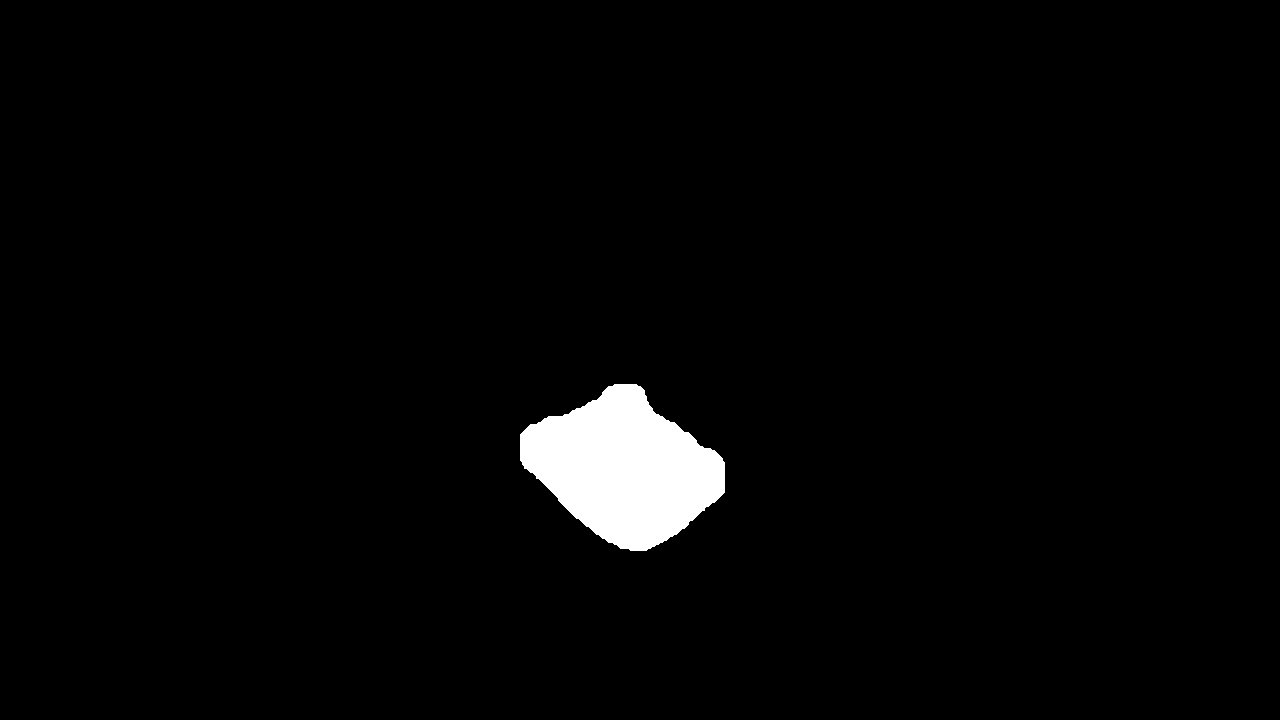}
        \caption{\label{fig:foot_yolo}}
    \end{subfigure}    
    \end{subfigure}
    \caption{System diagram of the footprint-estimation training process via the projection ray transform method (top) and excerpt of the footprint (bottom) as from the CAD ground truth (a), YOLO trained from the real-world ground truth (b) and from the synthetic ground truth (c).}
    \label{fig:footprint_pipeline}
    \vspace{-5pt}
\end{figure}

\subsection{Footprint Estimation with YOLO}
\label{sec: Footprint YOLO}

To show the validity of our approach, we train a YOLOv8 model on the labels created through the method described in \cref{sec:met:footprint}.
Here, we show two separate approaches, where we train YOLO on the real images used for training \gls{nerf} or fully synthetic ones.
We use 300 images in both cases and train a \texttt{yolov8m-seg} model for 100 epochs.
The resulting YOLO models predict the footprint of the 200 images in the evaluation dataset, which is separate from the training one.
A qualitative example can be seen in \cref{fig:footprint_pipeline}.

Calculating the \gls{ccr} of the predicted footprint against that found via ray projection, the first YOLO model achieves an average \gls{ccr} score of 0.849 but fails to segment a footprint in 14 out of the 200 images.
The second is always successful at segmenting a footprint while achieving a higher \gls{ccr} score of 0.873.
However, a qualitative survey of the footprints generated reveals that its predicted footprint tends to be blobbier, lacking the straight edges calculated from the projection ray transform method.
Critically for this application, though, blobbier images result in a conservative estimate, which may be favourable.

\section{Conclusion}

This work demonstrates a simple training procedure for extracting 3D priors to extend the operational areas of a \gls{ibvs} system. 
By estimating the relative pose of the fixed CCTV to the moving robot, we can construct its \gls{nerf} model through the CCTV footage with no additional priors.
This model can then extrapolate the robot's footprint and render a synthetic, labelled dataset to train a YOLO model to segment the robot's footprint from real data online at deployment time. 
We demonstrated its viability through simulation data, highlighting its simplicity and limitations in training data.

\bibliographystyle{IEEEtran}
\bibliography{biblio}

\begin{thebibliography}{10}
\providecommand{\url}[1]{#1}
\csname url@samestyle\endcsname
\providecommand{\newblock}{\relax}
\providecommand{\bibinfo}[2]{#2}
\providecommand{\BIBentrySTDinterwordspacing}{\spaceskip=0pt\relax}
\providecommand{\BIBentryALTinterwordstretchfactor}{4}
\providecommand{\BIBentryALTinterwordspacing}{\spaceskip=\fontdimen2\font plus
\BIBentryALTinterwordstretchfactor\fontdimen3\font minus \fontdimen4\font\relax}
\providecommand{\BIBforeignlanguage}[2]{{%
\expandafter\ifx\csname l@#1\endcsname\relax
\typeout{** WARNING: IEEEtran.bst: No hyphenation pattern has been}%
\typeout{** loaded for the language `#1'. Using the pattern for}%
\typeout{** the default language instead.}%
\else
\language=\csname l@#1\endcsname
\fi
#2}}
\providecommand{\BIBdecl}{\relax}
\BIBdecl

\bibitem{hill1979real}
J.~Hill, ``Real time control of a robot with a mobile camera,'' in \emph{Proc. 9th Int. Symp. on Industrial Robots}, 1979, pp. 233--245.

\bibitem{IBVS_1}
\BIBentryALTinterwordspacing
N.~Garcia-Aracil, C.~Perez-Vidal, J.~M. Sabater, R.~Morales, and F.~J. Badesa, ``Robust and cooperative image-based visual servoing system using a redundant architecture,'' \emph{Sensors}, vol.~11, no.~12, p. 11885–11900, Dec. 2011. [Online]. Available: \url{http://dx.doi.org/10.3390/s111211885}
\BIBentrySTDinterwordspacing

\bibitem{IBVS_2}
P.~Corke and S.~Hutchinson, ``A new partitioned approach to image-based visual servo control,'' \emph{IEEE Transactions on Robotics and Automation}, vol.~17, no.~4, pp. 507--515, 2001.

\bibitem{robinson2023robot}
L.~Robinson, M.~Gadd, P.~Newman, and D.~De~Martini, ``Robot-relay: Building-wide, calibration-less visual servoing with learned sensor handover network,'' \emph{arXiv preprint arXiv:2310.15677}, 2023.

\bibitem{robinson2023visual}
L.~Robinson, D.~De~Martini, M.~Gadd, and P.~Newman, ``Visual servoing on wheels: Robust robot orientation estimation in remote viewpoint control,'' in \emph{2023 IEEE/RSJ International Conference on Intelligent Robots and Systems (IROS)}.\hskip 1em plus 0.5em minus 0.4em\relax IEEE, 2023, pp. 6364--6370.

\bibitem{CAD}
J.~Feddema and O.~Mitchell, ``Vision-guided servoing with feature-based trajectory generation (for robots),'' \emph{IEEE Transactions on Robotics and Automation}, vol.~5, no.~5, pp. 691--700, 1989.

\bibitem{vanilla-nerf}
\BIBentryALTinterwordspacing
B.~Mildenhall, P.~P. Srinivasan, M.~Tancik, J.~T. Barron, R.~Ramamoorthi, and R.~Ng, ``Nerf: Representing scenes as neural radiance fields for view synthesis,'' 2020. [Online]. Available: \url{https://arxiv.org/abs/2003.08934}
\BIBentrySTDinterwordspacing

\bibitem{Yolo}
\BIBentryALTinterwordspacing
J.~Redmon, S.~Divvala, R.~Girshick, and A.~Farhadi, ``You only look once: Unified, real-time object detection,'' 2015. [Online]. Available: \url{https://arxiv.org/abs/1506.02640}
\BIBentrySTDinterwordspacing

\bibitem{sam}
A.~Kirillov, E.~Mintun, N.~Ravi, H.~Mao, C.~Rolland, L.~Gustafson, T.~Xiao, S.~Whitehead, A.~C. Berg, W.-Y. Lo, P.~Doll{\'a}r, and R.~Girshick, ``Segment anything,'' \emph{arXiv:2304.02643}, 2023.

\bibitem{wang2024dust3r}
S.~Wang, V.~Leroy, Y.~Cabon, B.~Chidlovskii, and J.~Revaud, ``Dust3r: Geometric 3d vision made easy,'' in \emph{Proceedings of the IEEE/CVF Conference on Computer Vision and Pattern Recognition}, 2024, pp. 20\,697--20\,709.

\bibitem{nerfstudio}
\BIBentryALTinterwordspacing
M.~Tancik, E.~Weber, E.~Ng, R.~Li, B.~Yi, J.~Kerr, T.~Wang, A.~Kristoffersen, J.~Austin, K.~Salahi, A.~Ahuja, D.~McAllister, and A.~Kanazawa, ``Nerfstudio: A modular framework for neural radiance field development,'' 2023. [Online]. Available: \url{https://arxiv.org/abs/2302.04264}
\BIBentrySTDinterwordspacing

\bibitem{inerf}
L.~Yen-Chen, P.~Florence, J.~T. Barron, A.~Rodriguez, P.~Isola, and T.-Y. Lin, ``{iNeRF}: Inverting neural radiance fields for pose estimation,'' in \emph{IEEE/RSJ International Conference on Intelligent Robots and Systems ({IROS})}, 2021.

\end{thebibliography}

\end{document}